# Using n-akṣaras to model Sanskrit & Sanskrit-adjacent texts

Presented at Perspectives of Digital Humanities in the Field of Buddhist Studies, Universität Hamburg, 13 January 2023. Revised 25 January 2023.

Charles Li

Centre nationale de la recherche scientifique

Paris, France

## Abstract

Despite — or perhaps because of — their simplicity, n-grams, or contiguous sequences of tokens, have been used with great success in computational linguistics since their introduction in the late $20^{\text{th}}$ century. Recast as k-mers, or contiguous sequences of monomers, they have also found applications in computational biology. When applied to the analysis of texts, n-grams usually take the form of sequences of words. But if we try to apply this model to the analysis of Sanskrit texts, we are faced with the arduous task of, firstly, resolving sandhi to split a phrase into words, and, secondly, splitting long compounds into their components. This paper presents a simpler method of tokenizing a Sanskrit text for n-grams, by using n-akṣaras, or contiguous sequences of akṣaras. This model reduces the need for sandhi resolution, making it much easier to use on raw text. It is also possible to use this model on Sanskrit-adjacent texts, e.g., a Tamil commentary on a Sanskrit text. As a test case, the commentaries on *Amarakoṣa* 1.0.1 have been modelled as n-akṣaras, showing patterns of text reuse across ten centuries and nine languages. Some initial observations are made concerning Buddhist commentarial practices.

---

This paper is an outcome of the Texts Surrounding Texts project of the CNRS, in collaboration with the Bibliothèque nationale de France, and jointly funded by the ANR & DFG (FRAL 2018).

# Introduction

N-grams were introduced in 1997, as a strategy for efficiently clustering documents on the Internet. Using the AltaVista search engine, researchers were able to perform syntactic clustering on "every document on the World Wide Web" at the time. In that paper, n-grams are called "shingles":

> We view each document as a sequence of words, and start by lexically analyzing it into a canonical sequence of tokens.... A contiguous subsequence contained in [a document] is called a *shingle*.[1]

The example given in the paper is the sentence "a rose is a rose is a rose." Split into 4-grams, or contiguous sequences of 4 words, this sentence yields the set "a rose is a," "rose is a rose," "is a rose is." This set can then be compared to other sentences, and the resemblance between them can be quantified by the difference between the sets.

This simple method of modelling a text — as subsequences of words — has been applied widely, not only on texts, but also, for example, on DNA. In fact, any phenomenon that has some extension in space and/or time can be modelled in this way.[2] The crucial question in each case is how to tokenize — how do you take a continuous reality and cast it into discrete units that will form the basis for n-grams?

# Where does meaning lie?

When we think of a text, we usually think of it as consisting of words, phrases, paragraphs, verses, etc. But it is important not to forget that every text also has a material basis — a text equally consists of incisions in a palm leaf, ink on a page, or sound waves propagating through air. These aspects, too, can be tokenized, and they can yield meaningful insights about a text and how it is transmitted — for example, in the case of palaeographic analysis or speech tone analysis.

---

1   Broder et al. 1997, 1158.
2   For a further example, see Huang et al., 2012, on n-grams used to cluster heartbeat signals.

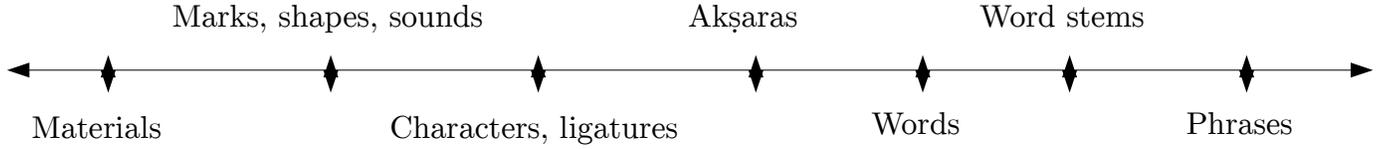

We can consider a text as a continuum — spanning its material basis, on the one end, and its linguistic or conceptual idealization, on the other. Focusing on any single point in the continuum implies some trade-off, some meaning lost and gained. For example, if we analyze a written text as words, with a normalized spelling, we ignore different orthographies that are used in different regions. On the other hand, if we analyze a text as a sequence of marks, perhaps for the purpose of handwriting analysis, we largely ignore its linguistic content. For Sanskrit texts, akṣaras are a good compromise between materiality and linguistics — they are a good representation of the sequence of marks on a written page while also giving some sense of semantics and syntax.

## Akṣaras vs. words

Akṣaras are well-defined and easy to tokenize. An akṣara is a single syllable ending in a vowel, anusvāra, or visarga. For example, the phrase *akṣaraḥ kartā* is tokenized as

<div align="center">a kṣa raḥ ka rtā</div>

There is no need for splitting words and compounds, and there is no need for stemming. This is especially important in cases where word splitting is undesirable, such as when the text is (intentionally) ambiguous, damaged or difficult to decipher. For example, take this manuscript fragment, where the commentary, written in the top and bottom margins, has been occluded by damage.

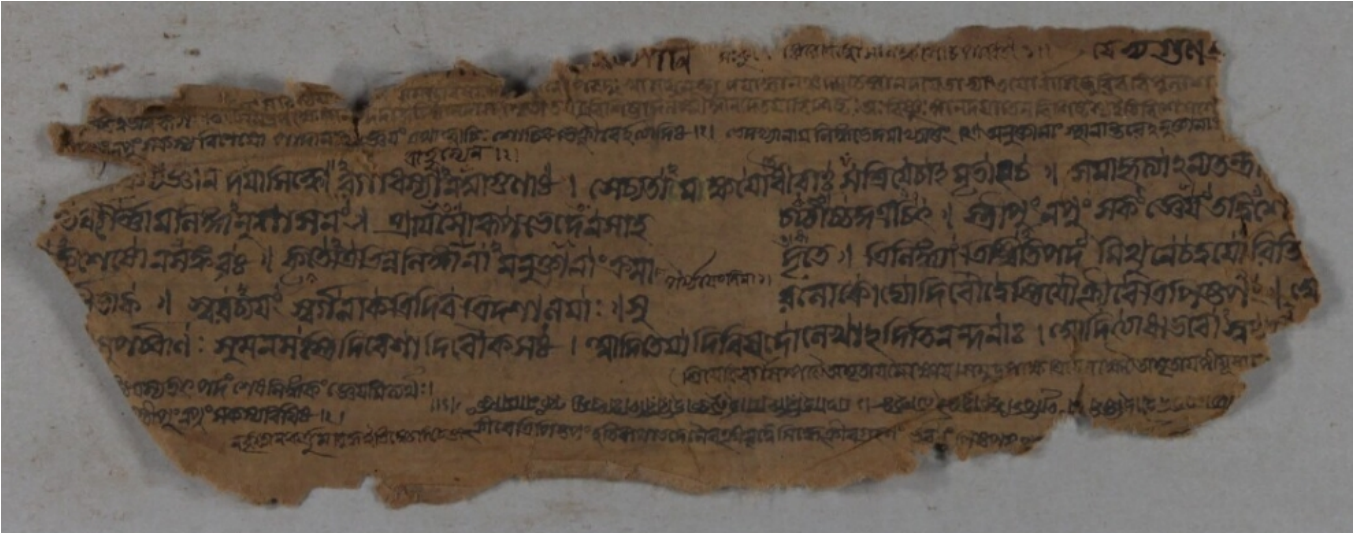

*Figure 1: Endangered Archives Project, Santipur Bangiya Puran Parishad. EAP781/1/1/1061, folio 1r.*

In a case such as this, where much of the text is lost, it can be difficult to determine where a phrase or even a word begins and ends. In order to split the text as words, it would be necessary to speculatively emend the text; if the emendation is wrong, then it will become impossible to match this text with similar passages in other manuscripts. Wrong emendations can also lead to compounding errors, bringing into being phantom words that may never have existed previously.[3] But if we tokenize this text as akṣaras, then we don't need to emend at all:

$$[...]\text{śritvabhāvaka}[...] \quad \rightarrow \quad \text{śri tva bhā va ka}$$

## Akṣaras vs. characters

It is possible to analyze akṣaras further into characters, i.e., as consonants and vowels. But n-grams composed of characters are less meaningful than n-grams composed of akṣaras. For example, compare these 2-grams of the previous example phrase, *akṣaraḥ kartā*, first as characters and then as akṣaras.

---

3    See Li 2022 for an example in Buddhist Hybrid Sanskrit.

$$\text{a kṣ a raḥ k a r t ā} \quad \rightarrow \quad \text{ak, kṣ, ṣa, ar, ra, aḥ, ḥk, ka, rt, tā}$$

$$\text{a kṣa raḥ ka rtā} \quad \rightarrow \quad \text{akṣa, kṣaraḥ, raḥka, kartā}$$

With characters, most of our n-grams don't convey much information; they are extremely common. In order to get meaningful results, we would need much longer sequences. But with akṣaras, even sequences composed of two tokens already capture single words or recognizable parts of words.

Moreover, akṣaras more closely match the written sequence of a text than characters do, even across the many different scripts used to write Sanskrit. For example, if we split the akṣara *kē* into (Romanized) characters, we get the sequence *k ē*. But in Indic writing systems, each consonant has an inherent vowel that needs to be suppressed with a virāma sign in order to express the consonant alone. And in many scripts, the vowel sign *ē* is written before the consonant *k*.

$$\text{Malayalam:} \quad \text{kē} \rightarrow \text{കേ} \quad \text{k ē} \rightarrow \text{ക് ഏ}$$

$$\text{Kannada:} \quad \text{kē} \rightarrow \text{ಕೇ} \quad \text{k ē} \rightarrow \text{ಕ್ ಏ}$$

By tokenizing a Sanskrit text as characters, we may end up misrepresenting the actual written sequence of the text, depending on the script used to express it.

# Commentaries on *Amarakoṣa* 1.0.1

As a test case, a corpus composed of commentaries on the first verse of the *Amarakoṣa* was modelled as n-akṣaras. The *Amarakoṣa,* or *Nāmaliṅgānuśāsana,* is a well-known Sanskrit lexicon by Amarasiṃha of uncertain date, but it is certainly the most widespread Sanskrit lexicon and perhaps the most widespread Sanskrit text in existence. It is also likely the most commented-upon Sanskrit text; it counts at least 80 known commentaries,[4] spanning from the 10th to the 21st centuries, not including anonymous and marginal commentaries. Perhaps one

---

4  Vogel 2015, 25.

reason why the text attracted so much commentary is the ambiguity of its first, benedictory verse:

> yasya jñānadayāsindhor agādhasyānaghā guṇāḥ |
> sevyatām akṣayo dhīrāḥ sa śriye cāmṛtāya ca || 1.0.1 ||

> Hey wise guys! For glory and for immortality, you should worship the one who is an unfathomable ocean of knowledge and compassion, whose qualities are faultless.

Amarasiṃha was almost certainly a Buddhist, and here, the one who should be worshipped almost certainly refers to the Buddha. But, because no deity is explicitly named, this verse has led to over a thousand years of commentarial speculation, interpretation, and hermeneutics:

> ihānukto 'pi buddho viśeṣaṇena spaṣṭaṃ pratīyate iti

> Here, even though unsaid, the Buddha is obviously understood through his qualities.
>
> <div style="text-align:right">Raghunātha Cakravartin (17th c.)</div>

> yady api śrīmadamarasiṃho buddhamatānuyāyī... tathāpi... śivasambandhivyākhyānaṃ naḥ sutarāṃ rocate

> Even if Amarasiṃha was a follower of Buddhism, still an explanation relating to Śiva would please us very much.
>
> <div style="text-align:right">Brahmānanda Tripāṭhin (20th c.)</div>

> svāmī tu jinam anusmṛtyeti... āha | tan na, jinavācakapadasyātrādarśanāt

> But [Kṣīra]svāmin said, "having memorialized the Jina." Not so, because a word expressing "Jina" does not appear in the verse.
>
> <div style="text-align:right">Bhānuji Dīkṣita (17th c.) quoting Kṣīrasvāmin (11th c.)</div>

> he dhīra saḥ aḥ viṣṇuḥ sevyatāṃ | akāro viṣṇuḥ

> Oh wise one! He, *aḥ*, or Viṣṇu, should be worshipped. *a* is Viṣṇu.
>
> anonymous marginal commentary, Shantipur Bangiya Puran Parishad MS A482

These commentaries are highly intertextual — they reuse and rephrase passages from one another, quote one another, and debate different interpretations of the verse. Many of them are also very inventive — the last excerpt, for example, divides the word *dhīrāḥ* into *dhīra* and *aḥ*. In fact, many commentators split the verse in different ways in order to obtain different meanings; they view the verse as a sequence of akṣaras that can be freely partitioned in different places. If we take the verse and split it into a canonical sequence of words, then that would only capture one out of its many possible meanings — we would be missing other interpretations of it as given by commentators over the centuries.

# Document similarity using n-akṣaras

By modelling the commentaries as n-akṣaras, we can quantify their similarity. For example, take these two phrases from two different commentaries:

> ihānukto 'pi buddho viśeṣaṇena spaṣṭaṃ pratīyate

> atrānukto pi budho viśeṣeṇaiḥ sūcayati

After normalizing the orthography — for features such as consonant gemination, e.g., *dho* vs. *ddho*[5] — we can compare these two phrases as sets of 4-akṣaras:

> ihānukto,  hānukto'pi,  **nukto'pibu**,  **kto'pibuddho**,  **pibuddhovi**,  **buddhoviśe**...

> atrānukto,  trānuktopi,  **nuktopibu**,  **ktopibudho**,  **pibudhovi**,  **budhoviśe**...

There are a number of similarity metrics that can then be used with these results; the Jaccard index,[6] for example, would quantify the similarity between these two sets as 0.5 (4 items in common / 8 unique items total).

---

5   On strategies for normalizing Sanskrit, see Li 2017.
6   Jaccard 1912.

# n-akṣaras across languages

Although the majority of commentaries on the *Amarakoṣa* are written in Sanskrit, many are written in other languages, such as Hindi, Newar, Tamil, etc. But since they often take inspiration from Sanskrit commentaries, or even quote them outright, there is a high degree of literal intertextuality between them that can be detected using n-akṣaras. For example, take these two passages from two commentaries in different languages:

| | | | |
|---|---|---|---|
| Marathi: | yasya jyā parameśvarāce | → | ya sya jyā **pa ra me śva** rā ce |
| Newar: | amo parameśvara | → | a mo **pa ra me śva** ra |

Here, we are able to find a 4-akṣara match since the same Sanskrit word has been used in both commentaries, even if it is inflected differently. But we can do better by implementing some fuzzy matching:

| | | | | | |
|---|---|---|---|---|---|
| Sanskrit: | śriye saṃpataye | → | śri ye saṃ pa ta ye | → | śri ye **saṃ pa t**a ye |
| | | | *normalization* | | *ignore n-2 vowels* |
| Malayalam: | śrī = sanpattŭ | → | śrī saṃ pa tŭ | → | śrī **saṃ pa t**ŭ |

Another strategy that we could employ is skip-grams,[7] i.e., we can skip akṣaras. This is especially useful because the languages we are working with feature inflectional suffixes; if we skip these suffixes, we can match sequences of word stems that are common across languages, without having to do any formal stemming:

| | |
|---|---|
| Hindi: | **sa**tya, **śau**ca, **da**yā, **kṣāṃ**ti, **tyā**ga ādi |
| | *skip 1 akṣara* |
| Sanskrit: | **sa**tyaṃ **śau**caṃ **da**yā **kṣāṃ**tiḥ **tyā**gaḥ |

---

7  Guthrie et al. 2006.

# Text reuse in *Amarakoṣa* commentaries

So far, 105 commentaries[8] on *Amarakoṣa* 1.0.1, in 9 languages, have been collected and modelled as sets of 2-, 3-, 4-, and 5-akṣaras. Using the Dice coefficient[9] as a measure of similarity, a minimum spanning tree was created, revealing patterns of text reuse.

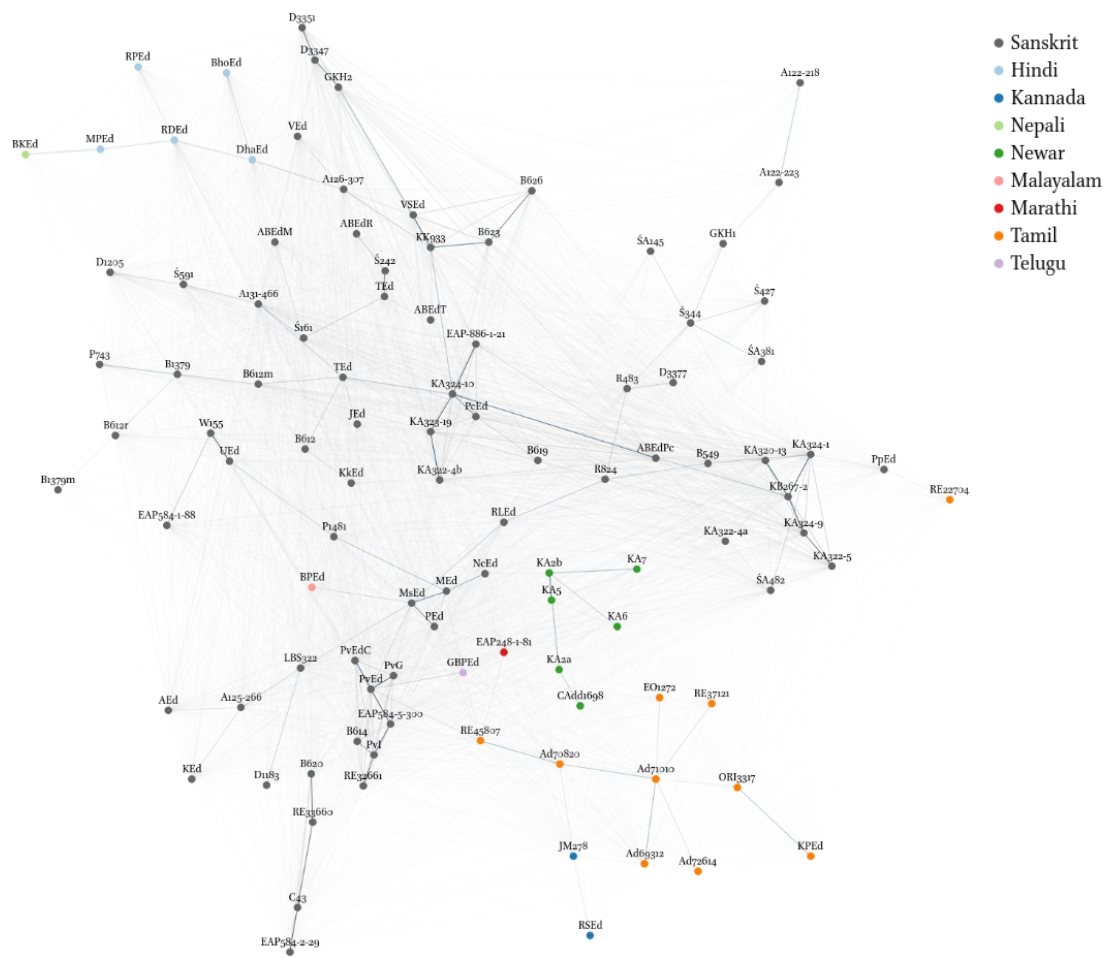

*Figure 2: Minimum spanning tree, using the Dice coefficient for edge weights. An interactive version of this figure can be found at* chchch.github.io/amarakosa

From this graph, some obvious clusters emerge — manuscripts of the same text are grouped very closely together. But the boundaries of these clusters are not always obvious. Especially in the case of anonymous commentaries, it is not always clear whether the intent was to copy an existing commentary or to create a new commentary based on older material. Scholars often distinguish between a text and witnesses of that text, as if there is one model which is simply copied imperfectly; but in the case of these commentaries, it can be very difficult to make that distinction, to know where one text ends and another begins. Perhaps a more accurate way to describe these clusters — rather than as witnesses of a distinct text — is as overlapping families of texts, without any particular centre considered as the urtext.

---

8   This count does not distinguish between "texts" and "witnesses"; even different versions of the same "text" are considered distinct texts themselves. See below.

9   Dice 1945.

Geographic regions and languages also create their own clusters — unsurprisingly, Tamil commentaries form a branch, connected to commentaries in other southern languages, while Hindi commentaries form a different branch along with the Nepali commentary, which seems to be a translation of a Hindi commentary. However, what seems at first to be a distinction between Sanskrit and non-Sanskrit commentaries may rather be a distinction between long, erudite commentaries and short, concise commentaries. While many of the Sanskrit commentaries are full of grammatical derivations and multiple interpretations, most of the commentaries in other languages are aimed at children, bearing titles such as *Bālapriyā* or *Bālabodhinī*, and they feature simple, common glosses.

But there are also many interesting outliers. RE22704, a Tamil and Telugu commentary preserved in a palm-leaf manuscript from the French Institute of Pondicherry, is conspicuously not connected to other Tamil commentaries but to PpEd, the *Padapārijāta*, a Sanskrit commentary. RE22704 is an unusually long and detailed commentary, and, in fact, it quotes the *Padapārijāta* — this is something that was already pointed out in a study of this manuscript by Giovanni Ciotti and R. Sathyanarayan,[10] and our quantitative analysis nicely mirrors their scholarly conclusions. The *Padapārijāta* itself is also difficult to place — although its author, Mallinātha, hails from south India, he quotes widely from both northern and southern Indian texts, brahmanical and Buddhist alike.[11]

---

10  Ciotti & Sathyanarayan 2020, 455-457.
11  Ramanathan 1971, xlvi-xlvii.

# Buddhisms

What is not apparent in the graph are clusters based on religion. Commentators did not limit themselves to only reading and quoting from authors of their same religious affiliation. Moreover, two commentators from what seems like the same religion may have radically different interpretations of the text. This is perhaps most evident when comparing two Buddhist commentaries: JEd, the 10th-century Sanskrit commentary of Jātarūpa, and KA2b, a Newar commentary preserved in a 16th-century manuscript in Kathmandu.[12]

Figure 3: Comparison of JEd and KA2b, with 3-akṣara matches highlighted. The online version of this figure can be reached at chchch.github.io/amarakosa: click the node JEd and then the node KA2b.

Both of these commentaries interpret the person who should be worshipped in *Amarakoṣa* 1.0.1 to be the Buddha, but the way in which the Buddha is described is completely different. Jātarūpa uses fairly technical language, describing the Buddha as *paramakāruṇika* and *vimalabuddhi*. On the other hand, KA2b uses the unusual phrase *parameśvara tathāgatatvaṃ*. These two commentaries have very little in common; at the level of 3-akṣaras, there are only a handful of matches, reflecting glosses that are common to many other commentaries. In fact, KA2b is less similar to JEd (Dice coefficient: 0.018) than it is to a Tamil Vaiṣṇava comment-ary that use *parameśvara* as an epithet of Viṣṇu (RE37121, Dice coefficient: 0.029). JEd and KA2b are each well-connected within their own geographical and cultural milieu — Jātarūpa, who may have been from Bengal, is quoted by other Bengali commentators, and KA2b is very similar to the other Newar commentaries — but they are practically unrelated to one another.

---

12  On JEd, see Pant 2000. On KA2b, see the description of A2b at newari.net/source.html.

## Going further

As has been suggested by other scholars at the conference, this study on text reuse in *Amarakoṣa* commentaries is confined to a small corpus, and it is a corpus in which a great deal of similarity between texts would already be expected. Further studies on a larger corpus may reveal surprising and unexpected connections between texts. In addition, a more formal comparison between character-, akṣara-, and word-level tokenization would be welcome, with performance metrics for different tasks and situations.

# Bibliography

British Library. 2014. "Nāmaliṅgānuśāsana or Amarakośa (with gloss)." Collection of manuscripts digitized from Shantipur Bangiya Puran Parishad, Endangered Archives Programme. doi:10.15130/EAP781 eap.bl.uk/archive-file/EAP781-1-1-1061

Broder, Andrei Z. et al. 1997. "Syntactic clustering of the Web." *Computer Networks and ISDN Systems* 29: 1157-1166.

Ciotti, Giovanni & Sathyanarayan, R. 2020. "A multilingual commentary of the first verse of the *Nāmaliṅgānuśāsana* as found in ms. IFP RE22704." In Anandakichenin, S. & D'Avella, V., eds., *The Commentary Idioms of the Tamil Learned Traditions*: 443-489. Pondichéry: École française d'Extrême-Orient.

Dice, Lee R. 1945. "Measures of the Amount of Ecologic Association Between Species." *Ecology* 26(3): 297–302. doi:10.2307/1932409

Guthrie, David et al. 2006. "A Closer Look at Skip-gram Modelling." *Proceedings of the Fifth International Conference on Language Resources and Evaluation (LREC'06), Genoa, Italy*: 1222-1225.

Huang, Yu-Chen et al. 2012. "Using n-gram analysis to cluster heartbeat signals." *BMC Medical Informatics and Decision Making* 12(64). doi:10.1186/1472-6947-12-64

Jaccard, Paul. 1912. "The Distribution of the Flora in the Alpine Zone." *New Phytologist* 11(2): 37–50. doi:10.1111/j.1469-8137.1912.tb05611.x

Li, Charles. 2017. "Critical diplomatic editing: Applying text-critical principles as algorithms." *Advances in Digital Scholarly Editing,* ed. P. Boot et al. Leiden: Sidestone Press.

Li, Charles. 2022. "Akālaka: a lexical phantom in Buddhist Hybrid Sanskrit." *The World of the Orient* 4: 203-210. doi:10.15407/orientw2022.04.203

Nepal Bhasha Dictionary Committee. "Description of Source manuscripts of Amarakośas." Newari Lexicon based on the Amarakosa. newari.net/source.html

Pant, Mahes Raj. 2000. *Jātarūpa's Commentary on the Amarakoṣa.* Delhi: Motilal Banarsidass.

Ramanathan, A. A. 1971. *Amarakośa [1] with the unpublished south Indian commentaries.* Madras: Adyar Library and Research Centre.

Vogel, Claus. 2015. *Indian Lexicography. Revised and Enlarged edition.* München: P. Kirchheim.